\documentclass[conference]{IEEEtran}
\IEEEoverridecommandlockouts
\usepackage{cite}
\usepackage{amsmath,amssymb,amsfonts}
\usepackage{algorithmic}
\usepackage{graphicx}
\usepackage{amsmath}
\usepackage{amssymb}
\usepackage[utf8]{inputenc}
\usepackage{float}
\usepackage{caption}
\usepackage{subcaption}
\usepackage{tabularx}
\usepackage{booktabs} 
\usepackage{textcomp}
\usepackage{xcolor}
\def\BibTeX{{\rm B\kern-.05em{\sc i\kern-.025em b}\kern-.08em
    T\kern-.1667em\lower.7ex\hbox{E}\kern-.125emX}}
  
\def \paperPath {}
    
\begin{document}

\title{Distance in Latent Space as Novelty Measure\\
}

\author{\IEEEauthorblockN{1\textsuperscript{st} Mark P. Philipsen}
\IEEEauthorblockA{\textit{Danish Meat Research Institute} \\
\textit{Danish Technological Institute}\\
Taastrup, Denmark \\
mpph@create.aau.dk}
\and
\IEEEauthorblockN{2\textsuperscript{nd} Thomas B. Moeslund}
\IEEEauthorblockA{\textit{Visual Analysis of People Group} \\
\textit{Aalborg University}\\
Aalborg, Denmark \\
tbm@create.aau.dk}
}

\maketitle

\begin{abstract}
Deep Learning performs well when training data densely covers the experience space. For complex problems this makes data collection prohibitively expensive.
We propose to intelligently select samples when constructing data sets in order to best utilize the available labeling budget. The selection methodology is based on the presumption that two dissimilar samples are worth more than two similar samples in a data set.
Similarity is measured based on the Euclidean distance between samples in the latent space produced by a DNN. By using a self-supervised method to construct the latent space, it is ensured that the space fits the data well and that any upfront labeling effort can be avoided.
The result is more efficient, diverse, and balanced data set, which produce equal or superior results with fewer labeled examples.
\end{abstract}

\begin{IEEEkeywords}
Deep Learning; Active Learning, Data sets
\end{IEEEkeywords}

\section{Introduction}
Data set collection and preparation is likely to take up a majority of the time and effort applying Machine Learning in real-world application. The high upfront cost of acquiring large amounts of quality labeled data makes it difficult to gather sufficient samples to cover the possible variation and properly evaluate the robustness of a solution. With complex problems where labeling is more involved than simple categorization it is difficult to reach the data scales necessary for building truly robust learning based systems. Medical imaging~\cite{kohli2017medical} and related areas are especially challenged in this regard because of the ambiguity in labeling. Labeling may require expert knowledge and even then the ambiguities are likely to lead to inconsistent results. The difficulty and expense of labeling means that it is highly desirable to be economical w.r.t. selecting the samples that improve the model the most.
The attempt to intelligently select a subset of samples from a larger data set is called Active Learning (AL) and has been studied extensively. The problem has been addressed in a number of ways such as ensemble-based methods~\cite{beluch2018power}, Monte-Carlo Dropout~\cite{gal2016dropout} and geometric approaches~\cite{sener2017active}. However, many classical strategies and heuristics are not effective for DNNs, partly because DNNs are arbitrarily confident when mistaken, but mainly because these methods do not cover large parts of the sample space~\cite{sener2017geometric}.

The root of the problem is the imbalances that exists within data sets due to natural differences in the frequencies with which different classes occur and that each sample isn't equally valuable to the learner. Imbalanced data sets are typically considered a problem in multi-class problems where each class is not equally represented. In applications such as disease classification and other types of abnormality detection it is common that one class, the majority class, contains a lot more samples compared to minority classes~\cite{johnson2019survey}. In these cases, learners will typically overly focus on the majority group which dominate the gradients that update the network parameters~\cite{anand1993improved}.
In-class imbalances result in similar problems but are much harder to mitigate. Imbalanced multi-class problems can be fairly evaluated using confusion matrices and class weighted metrics. Learners can balance gradient contributions across classes in proportion to the number of samples in each class~\cite{cui2019class}. With in-class imbalances it is difficult to quantify the severity of the imbalance and there are no clear measures compensate for it. The problem of in-class imbalances is generally much less explored.

Solutions for dealing with class imbalances include~\cite{DBLP:journals/corr/abs-1807-04950}; down-sampling the majority class, up-sampling the minority class e.g using image augmentation, transfer learning, increasing the weight of the minority class in the optimization loss, and generating synthetic data for the minority class using techniques such as SMOTE~\cite{DBLP:journals/corr/abs-1106-1813}, Cavity Filling~\cite{DBLP:journals/corr/abs-1807-06538} or GANs~\cite{DBLP:journals/corr/abs-1810-10863,antoniou2017data}. Method that generate synthetic data through interpolation~\cite{DBLP:journals/corr/abs-1106-1813,he2008adasyn} or sampling from a learned distribution~\cite{DBLP:journals/corr/abs-1807-06538,antoniou2017data} suffer the risk of generating false positives and may exacerbate in-class imbalance.
Here, we take the approach of effectively down-sampling the dense areas in feature space of a DNN. This is a viable option when large amounts of raw data is available and labeling is the limiting factor.
The work originates in a need to optimize the labeling effort such that the samples that are selected for use are the samples that provide the best basis for training and evaluation, avoiding to spend labeling effort on unnecessary near duplicate samples.






\subsection{Contributions}
This paper addresses a key concern that show up when applying ML. Namely, constructing cost effective training sets. The contributions are summarized here:
\begin{enumerate}
    \item Reduce labeling effort through selection of training samples informed by distances in feature space. 
    \item Balance data sets suffering from in-class imbalances.
\end{enumerate}

\section{Related Work}
The high cost of acquiring large amounts of quality labeled data makes preparation of comprehensive data sets expensive. The aim of the active learning field is to reduce the labeling effort and optimized learning either lead by an algorithm which queries a human for labels, or by a human who decides which samples to label~\cite{thomaz2006reinforcement}.
The majority of published work on active learning takes the algorithm route. A typical example is to start with a model trained on a small subset of the data set and iteratively add more samples based on the models confidence on the remaining unlabeled samples~\cite{aghdam2019active}. The traditional approach is inverted in~\cite{wang2016cost} where a model again is trained in an incremental fashion but noticeably by selecting high confidence unlabeled samples for their use in feature learning~\cite{wang2016cost}. 
A geometric perspective is taken in~\cite{sener2017geometric}, namely by hypothesizing that the labeled samples should cover the large unlabeled data set as closely as possible. This is achieved by minimizing a core-set loss, which is the difference between the loss for the selection and the entire data set.


The literature on combating class imbalances is generally addressing the problem from a data generation approach.
Synthetic Minority Over-sampling Technique (SMOTE) generates new observations for minority classes by interpolating between samples in the original data set. Their method effectively over-sample the minority class and under-sample the majority class. They propose different rules for how to handle the generated samples. They achieve better performance than simply under-sampling the majority class~\cite{DBLP:journals/corr/abs-1106-1813}. Adaptive Synthetic Sampling (ADASYN) functions similarly to SMOTE, except that focus is placed on generating synthetic training examples near outliers, adjusting the number of samples generated proportionally to the number of nearby samples belong a different class~\cite{he2008adasyn}. "Cavity filling" is a technique which generates pseudo-features for filling the gaps between the minority and majority classes in feature spaces. Pseudo-features are generated from multivariate probability distributions from the features of minority classes Features are extracted from a layer of trained deep neural network. Notably they do not generate pseudo-data but pseudo-features. The pseudo-features in combination with real features are used to train the layers following the layer which the features belong~\cite{DBLP:journals/corr/abs-1807-06538}.



\section{Method}
\begin{figure}[h]
\centering
\begin{subfigure}{0.49\columnwidth}
  \centering
  \includegraphics[width=1.0\linewidth]{\paperPath 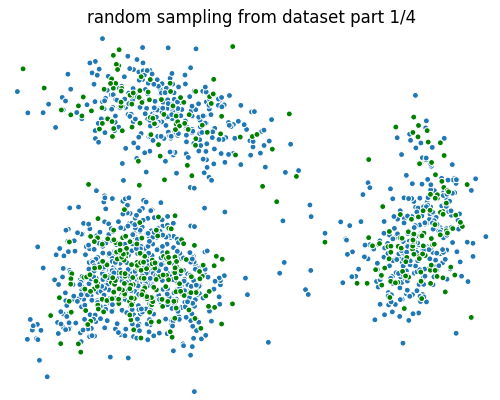}
  \caption{}
  \label{method:selection_from_distance_vs_random:fig:ra14}
\end{subfigure}
\hfill
\begin{subfigure}{0.49\columnwidth}
  \centering
  \includegraphics[width=1.0\linewidth]{\paperPath 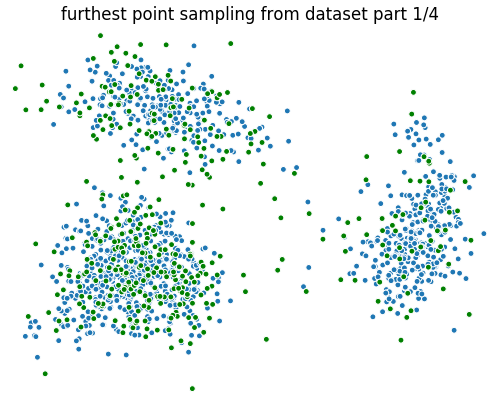}
  \caption{}
  \label{method:selection_from_distance_vs_random:fig:fu14}
\end{subfigure}
\newline
\begin{subfigure}{0.49\columnwidth}
  \centering
  \includegraphics[width=1.0\linewidth]{\paperPath 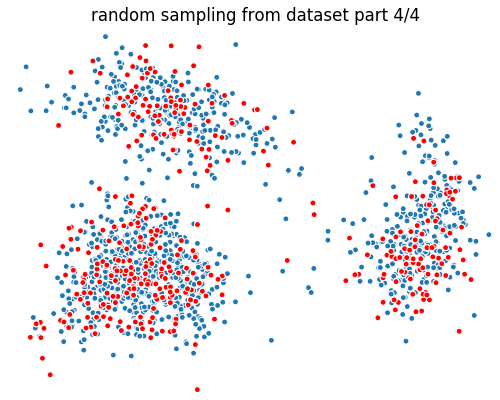}
  \caption{}
  \label{method:selection_from_distance_vs_random:fig:ra44}
\end{subfigure}
\hfill
\begin{subfigure}{0.49\columnwidth}
  \centering
  \includegraphics[width=1.0\linewidth]{\paperPath 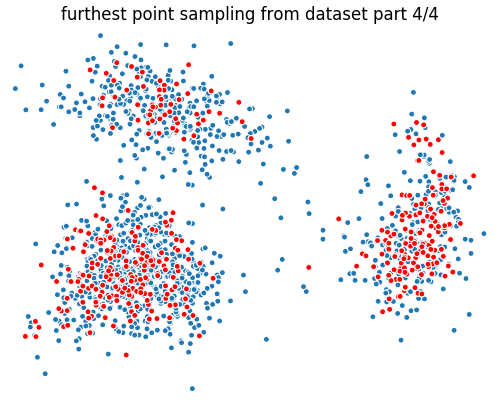}
  \caption{}
  \label{method:selection_from_distance_vs_random:fig:fu44}
\end{subfigure}
\caption{(a) First $1/4$ sample using random sampling. (b) First $1/4$ sample using furthest point sampling. (c) Last $1/4$ sample using random sampling. (d) Last $1/4$ sample using furthest point sampling. (blue) Unselected samples, (green) First $1/4$ samples selected, and (red) Last $1/4$ samples selected.}
\label{method:selection_from_distance_vs_random}
\end{figure}

The intuition behind the proposed method is that the known distribution should be uniformly distributed in latent space in order to effectively minimize the distance between new samples and known samples under the constraint of a limited number of known samples. This can be done by relying on learned feature space representation. A self-supervised method for learning representations is shown in~\cite{DBLP:journals/corr/LarsenSW15}, but representations can be extracted from any relevant DNN. In this case an Autoencoder is used to build the feature space transformation in a self-supervised manner. For visualization purposes the dimensionality of the latent representations is reduced using Principal component analysis and only the two foremost principal components are used in the following plots.  Figure~\ref{method:selection_from_distance_vs_random} shows a comparison between the samples selected using random sampling and sampling maximizing distances in feature space. Figure~\ref{method:selection_from_distance_vs_random:fig:ra14} and~\ref{method:selection_from_distance_vs_random:fig:fu14} shows the first $1/4$ of the total number of samples, sampled using the two different methods. Figure~\ref{method:selection_from_distance_vs_random:fig:fu14} shows a more uniformly sample (green) from the data set(blue) using furthest point sampling. Training using this sample should provide better coverage of the input space compared to the random sample shown in Figure~\ref{method:selection_from_distance_vs_random:fig:ra14}.
Figure~\ref{method:selection_from_distance_vs_random:fig:ra44} and~\ref{method:selection_from_distance_vs_random:fig:fu44} shows the last $1/4$ of the selected samples. The random sample shown in Figure~\ref{method:selection_from_distance_vs_random:fig:ra44} is clearly better distributed in the feature space compared to the sample using furthest point sampling shown in Figure~\ref{method:selection_from_distance_vs_random:fig:fu44}. The result is that furthest point sampling provides the most significant samples up front, while each batch selected using random sampling is approximately equally diverse.

\begin{figure}[t]
  \centering
  \includegraphics[width=0.8\columnwidth]{\paperPath 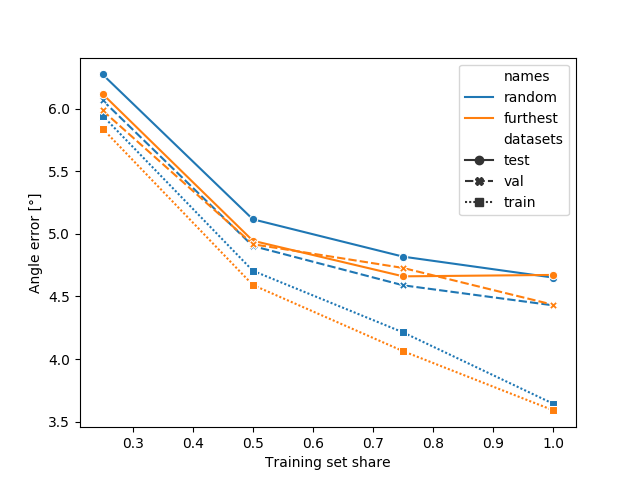}
  \caption{Prediction performance in relation to training set size.}
  \label{method:sampling_results}
\end{figure}

The two sampling approaches are compared by training models using the selected samples. In this case the model must predict the orientation of an object. Figure~\ref{method:sampling_results} shows that the model that is trained on fractions of the data set selected using furthest point sampling outperform random sampling, with the addition of each batch until all of the data is in use. 


\section{Discussion}
The preliminary results from applying the proposed method for selecting samples in latent space indicate that it is worth being selective w.r.t the subset of samples that are labeled. It is likely that the effects will be more pronounced with larger data sets, where the number of near duplicates can be expected to be exponentially larger.
Active learning methods are valuable to the development and deployment of models for real-world applications but see limited use. The reason for the limited success of these methods is likely found in their complexity and is less likely their overall effectiveness. Active learning methods should thus be evaluated based on other parameters than absolute reduction in labeled examples.

\section*{Acknowledgment}
This work was supported by Innovation Fund Denmark and the Danish Pig Levy Fund.

\bibliographystyle{IEEEtran}
\bibliography{main}

\end{document}